\documentclass[final,5p,times]{elsarticle}
\usepackage{amssymb}
\usepackage{graphicx}
\usepackage{float}
\usepackage[T1]{fontenc}
\usepackage{amsmath,color}
\usepackage{hyperref}

\journal{Computers \& Graphics}

\begin{document}

\begin{frontmatter}

\title{\textbf{Digital Makeup from Internet Images }}

\author{Asad Khan$^{1}$, Muhammad Ahmad$^{2}$, Yudong Guo$^{1}$, Ligang Liu$^{1}$}

\address{$^{1}$Graphics and Geometric Computing Lab, School of Mathematical Sciences,\\
University of Science and Technology of China,\\
Hefei, Anhui, 230026, PR China}
\address{$^{2}$Machine learning and knowledge representative(MIKr),\\
Innopolis University,\\
Innopolis,Russia}

\begin{abstract}
We present a novel approach of creating face
makeup upon a face image with another images as the style
examples. Our approach is analogous to physical makeup,
as we modify the color and skin details while preserving the
face structure. More precisely, we extract image foregrounds from both subject and multiple example images.
Then by using image matting algorithms, the system extracts the semantic information such as
faces, lips, teeth, eyes, eyebrows, etc., from the extracted foregrounds of both subject and multiple example images.
And, then the makeup style is transferred between the corresponding parts with the same semantic information.
Next we get the face makeup transferred result by seamlessly compositing different parts
together using alpha blending. In the final step, we present an efficient method of makeup consistency
to optimize the color of a collection of images showing the common scene.
The main advantage of our method over existing
techniques is that it does not need face matching, as one could use more than one example images.
Because one example image does not fulfill the complete requirements of a user.
Our algorithm is not restricted to head shot images as we can also change the makeup style in the wild.
Moreover, our algorithm does not require to choose the same pose and image size between subject
and example images. The experiment
results demonstrate the effectiveness of the proposed
method in faithfully transferring makeup.

\end{abstract}


\end{frontmatter}

\section{Introduction}

It has always been of special interest to humans to improve looks which can be
misleading sometimes. There are increasingly many commercial facial makeup systems in the
market, as makeup makes an individual more attractive and beautiful.
Face makeup can be described as a technique to change an individual's
appearance by using special cosmetics such as lipstick, foundation, powders,
creams etc. It is commonly used among females to enhance attraction in their
natural appearances. The foundation and loose powder are commonly used
to change the texture of face's skin, while applying makeup physically.
The first step is usually to use the foundation to conceal flaws and cover
the original skin texture, and then loose powder is mainly used to introduce
new, usually eye-catching and pleasant textures to skin. The application of other
makeup constitutes like rouge, eye liner and shadow, etc., follow on top layer of
the powder. Nevertheless, color makeup transfer is still a challenging task as almost
all of the existing techniques undergo with a number of technical and computational disadvantages.
Thus it has been a contemporary focus of research to successfully transfer the color makeup in
digital images.\\

After entering in a beauty salon, the customer usually selects
\begin{figure}[t!]
  \centering
  \includegraphics[width=3.5in]{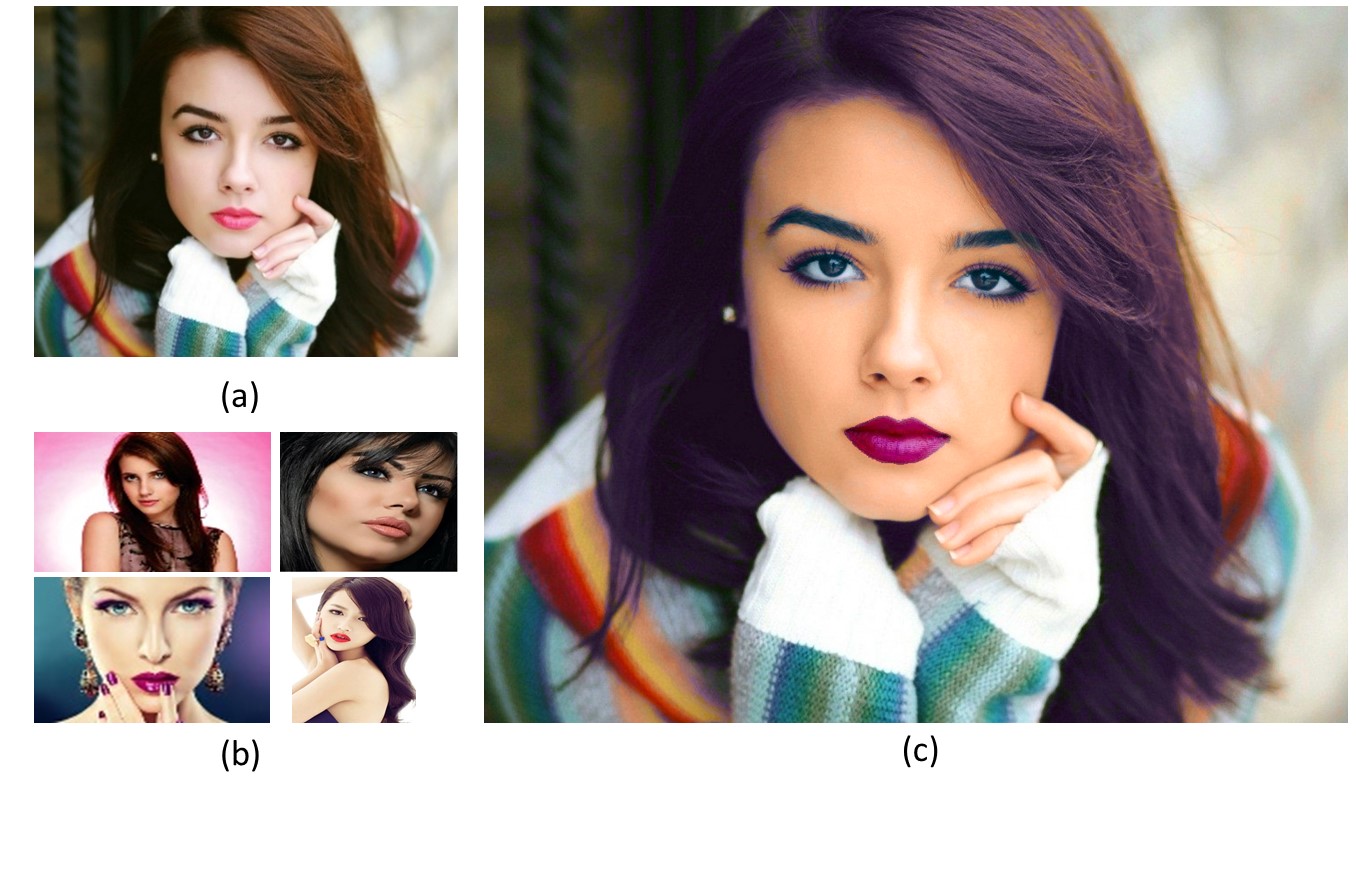}\vspace{-0.5cm}
  \caption{Digital makeup from internet images. (a) A subject image, taken
by a common user. (b) example style images, download from internet. (c) The result of our approach,
where foundation effect, eye style, eyebrow style, hair style, skin style and lip highlight in (b) are
successfully transferred to (a).}
  \label{fig:image1}
\end{figure}
an artistic example image from an available catalog of example images
and tells the makeup artist to apply the same makeup on her.
It would be more convenient for the user to select more than one example images
of her choices from the examples catalog as it would allow her to choose different parts
of the face from different example images.
Before actually applying the selected task, it would be quit helpful
if she is able to preview the same makeup style to her own face.
However, it is much time-consuming. Occasionally, the customer has two choices
for trying out makeup, of which one is to try the makeup physically which
is much time-consuming and requires the customer to be patient.
One of the possible way is to try the makeup digitally by ways of digital
photography by using different photo editing environments, such as
Adobe Photoshop TM \cite{25}. Besides,
\begin{figure*}
  \centering
  \includegraphics[width=\linewidth]{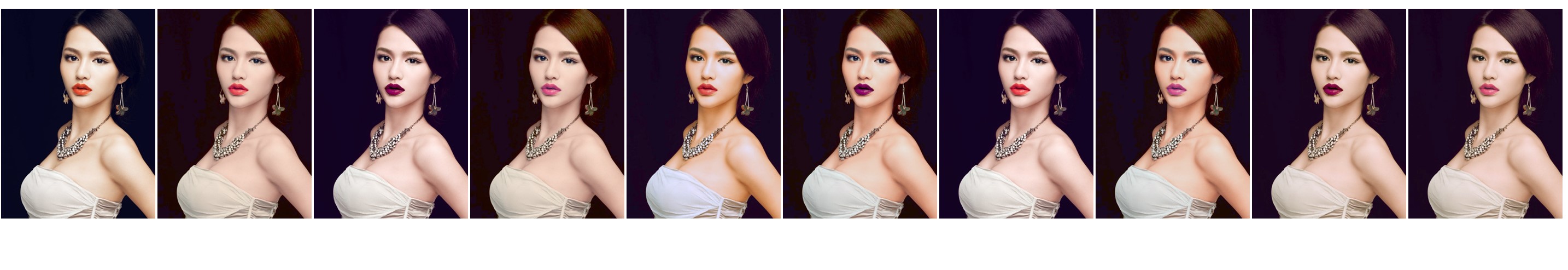}\vspace{-0.5cm}
  \caption{Some results with different facial styles by considering a single subject image and a number of multiple example images.}
  \label{fig:different style}
\end{figure*}
an online commercial software, Taaz \cite{24}, provides
users with virtual makeup on face photos by simulating the
effects of specified cosmetics. But then using such type of method is usually tedious
and relies heavily on the personal expertise and efforts of the user.\\

This paper deals with the problem of creating makeup upon
a face image ( Fig. \ref{fig:image1}(a))with the prototype of multiple images ( Fig. \ref{fig:image1}(b))as the style examples.
This is very practical while applying it in the scenario of the beauty salon.\\

Our approach is inspired by the process of physical
makeup. We present an approach of creating makeup
upon a face image with the prototype of another
image as the style example.
In transferring face makeup between images, we have the following challenges. First, such a technique
must maintain the correspondences between meaningful image regions in an automatic way. Secondly,
for novice users, the pipeline should be intuitive and user friendly. Thirdly, an efficient technique
to optimize makeup consistency of a collection of images depicting a common
scene. The generation of automatic Trimap is another challenge as almost
all of the existing techniques require a user to input a Trimap
manually.\\

In this paper, we propose a approach which is effective in
transferring face makeup from all types of images by taking advantage of high level
facial semantic information and large-scale Internet photos by professional
artists. A user can retouch his image easily to achieve a compelling
visual style by using such an algorithm. We present a matrix factorization based
approach to automatically optimize makeup consistency for
multiple images using sparse correspondence obtained from
multi-image sparse local feature matching. For rigid scenes,
we leveraging structure from motion (SfM) although it is an
optional step. We stack the aligned pixel intensities into a
vector whose size equals the number of images. Such vectors
are stacked into a matrix, one with many missing entries.
This is the observation matrix that will be factorized.
Under a simple makeup correction model, the logarithm of this
matrix satisfies a rank two constraint under idealized conditions.\\

In short summary, this article makes the following contributions:\\

$\bullet$ a new face makeup transferring technique is presented which
can transfer makeup between different regions of the subject image and multiple example images with the same facial
semantic information,\\

$\bullet$ we propose a new algorithm of automatic generation of Trimap for efficient synthesis of each facial semantic
information,\\

$\bullet$ a semantic makeup style transfer technique which transfers the makeup automatically is presented and an efficient technique
to optimize makeup consistency of a collection of images depicting a common
scene.\\

More importantly, our proposed method does not require the user to choose the same pose, face matching and
image size between subject and example images.
We demonstrate high quality makeup
consistency results on large photo collections of internet images.


\section{Related Work}

Not much work on digital makeup transfer has been done so far.
The first attempt for the makeup transfer was made by Guo et al. \cite{1}.
It first decomposes the subject image and example image faces into three
layers. Then, they transfer information following the layer by layer correspondence.
It is considered to be a disadvantage that it requires warp the example image to the
subject face, which seems very challenging. The work of Scherbaum et al. \cite{2} makes
use of a 3D morphablle face model to conduct facial makeup. The main disadvantage to their
technique is that it requires the before-after makeup face pairs of the same individual, which
is difficult to accumulate in real application. Tong et al. \cite{3} propose a "cosmetic transfer"
procedure to realistically transfer the cosmetic style
captured in the example-pair to another person's face.
The limitation of the applicability of the system occurs when requiring subject and example image face
pairs. The automatic makeup recommendation and synthesis system proposed by Liu et al. \cite{4} is another
work on digital makeup. They called their recommendation beauty e-expert. But their proposed technique is
considered to be in the recommendation module. By summing up this discussion, our proposed technique enhances
the applicability and is more flexible to the requirements of digital makeup methods, thus consequently
generates more attractive and eye-catching results.\\

\begin{figure}
  \centering
  \includegraphics[width=\linewidth]{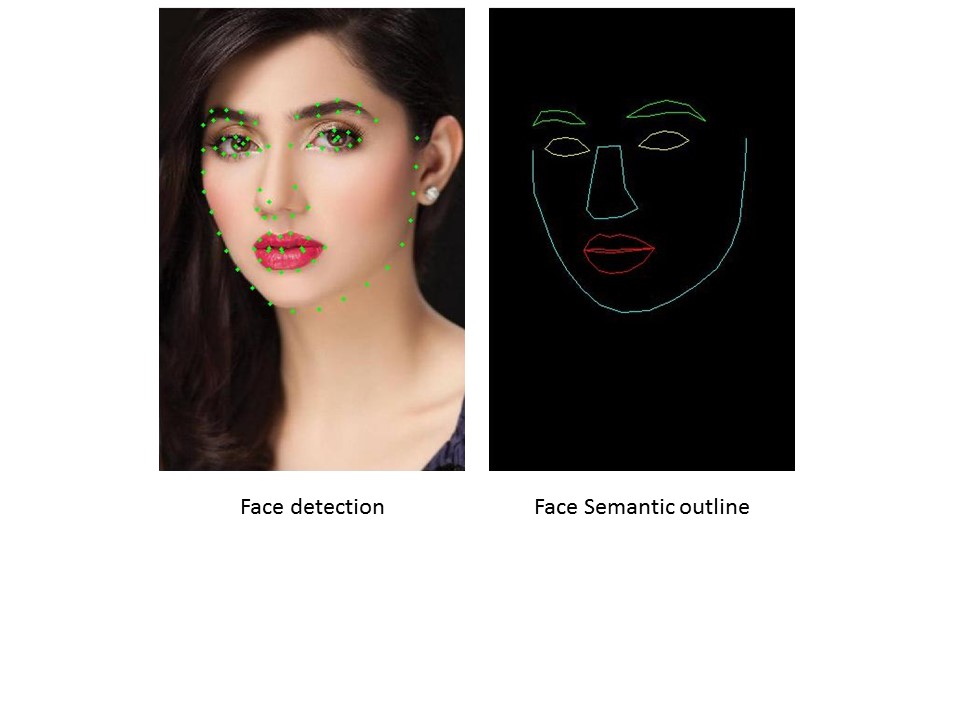}\vspace{-1.8cm}
  \caption{we connect the 83 key points on the face based on face detection, then the facial
semantic information outline can be obtained.}
  \label{face outline}
\end{figure}

There is another method proposed by Ojima et al. \cite{5}. It also
use the concept of subject-example images. But they have only discussed
the foundation effect in their work. If we study the contrast, the work
by Tsumura et al. \cite{6} proposed a physical model to extract
hemoglobin and melanin components. They simulated the changes in physical
facial appearance by adjusting the quantified amount of hemoglobin and melanin.
They addressed the effects including tanning and reddening due to cosmetic effects,
aging and alcohol consumption. Whereas the cosmetic effects they have used are much limited,
and demonstrate much simplification than the real makeup. Nevertheless, there is an online
commercial software called, Taaz \cite{24}, which accommodates the users while simulating the
effects of particular cosmetics and provides a virtual makeup on facial photos.\\

Recently, there is a subsequently successful fashion analysis work obtained by deep learning \cite{7,8}.
There is a recent work on this topic by Dosovitskiy et al. \cite{9,10}. They generate images of different objects
of given type, viewpoint and color by using a generative CNN. The work by Simonyan et al. \cite{11} generates an image
which captures a net and visualizes the notion of the class. There is a general framework to invert both hand-crafted and
deep representations to the image provided by Mahendran et al. \cite{12}.
Gatys et al. \cite{13} contribute a neural algorithm of artistic style based on the CNN.
Goodfellow et al. \cite{14} presented a generative network of adverse nature and comprises two components;
a generator and a discriminator. Without using obvious artifacts, the generated image is much neutral.
A much simpler and relatively faster method of generating adversarial example image was provided by Goodfellow et al. \cite{15}.
They mainly focus on enhancing the CNN training instead of image synthesis. The so-called Deep Convolution Inverse Graphics Network
proposed by Kulkarni et al. \cite{16} learns an interpretable representation of images for 3D rendering.
There is a considerable disadvantage by all existing deep methods that they generate only single image.
Apart from these facts, we mainly focus on how to generate a new image comprising the nature of two subject images.\\

\begin{figure}
  \centering
  \includegraphics[width=\linewidth]{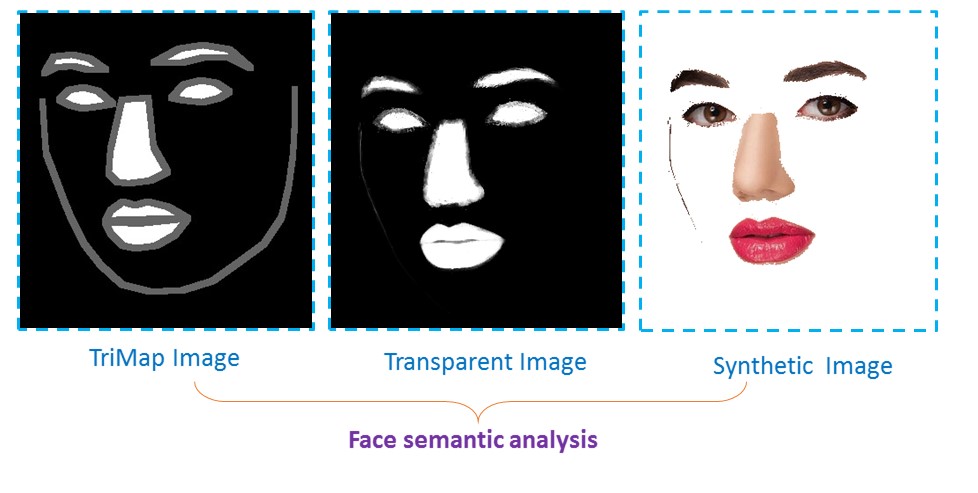}\vspace{-0.3cm}
  \caption{Trimap in the leftmost image while transparent image and synthetic image are the medal and right respectively.}
  \label{fig:trimap}
\end{figure}


\section{Digital Makeup}
\noindent\textbf{3.1.    Face Database}\\

There are lots of attractive and artistic images on the Internet. These images are produced
by professional photographers and professional cameras. It would be interesting if
ordinary people could reproduce the styles of these photos using some simple image
processing operations. Therefore, we use a database to store example images and their semantic information.
All the images in the database are segmented using matting techniques. we detect the key points of a human face and obtain
the facial characteristics. In this article, we utilize the API provided by the \cite{21,26} for
face detection. The landmark API can detect the key points of a human face robustly. The
API is used to detect the position of the facial contours, facial features and other key points.
Our approach detects $83$ key points in the face are depicted
in Fig. \ref{face outline}.\\

\noindent\textbf{3.2.    Contours of Face Semantics}\\

We can focus on the human face semantic analysis. In a certain order, we connect the $83$ key points
on the face based on face detection, then the facial semantic information outline can be obtained.
The landmark entry stores the key points of the
various parts of the human face, including eyebrows, eyes, mouth, chin, face and so on. Each
part has some points, and the points are represented by the coordinates using $x$ and $y$. Using
these key points, we connect them in a certain order and then we get the contour of the
face. The face semantic information
outline is given in Fig. \ref{face outline}.\\

\begin{figure*}
  \centering
  \includegraphics[width=18cm,height=9cm]{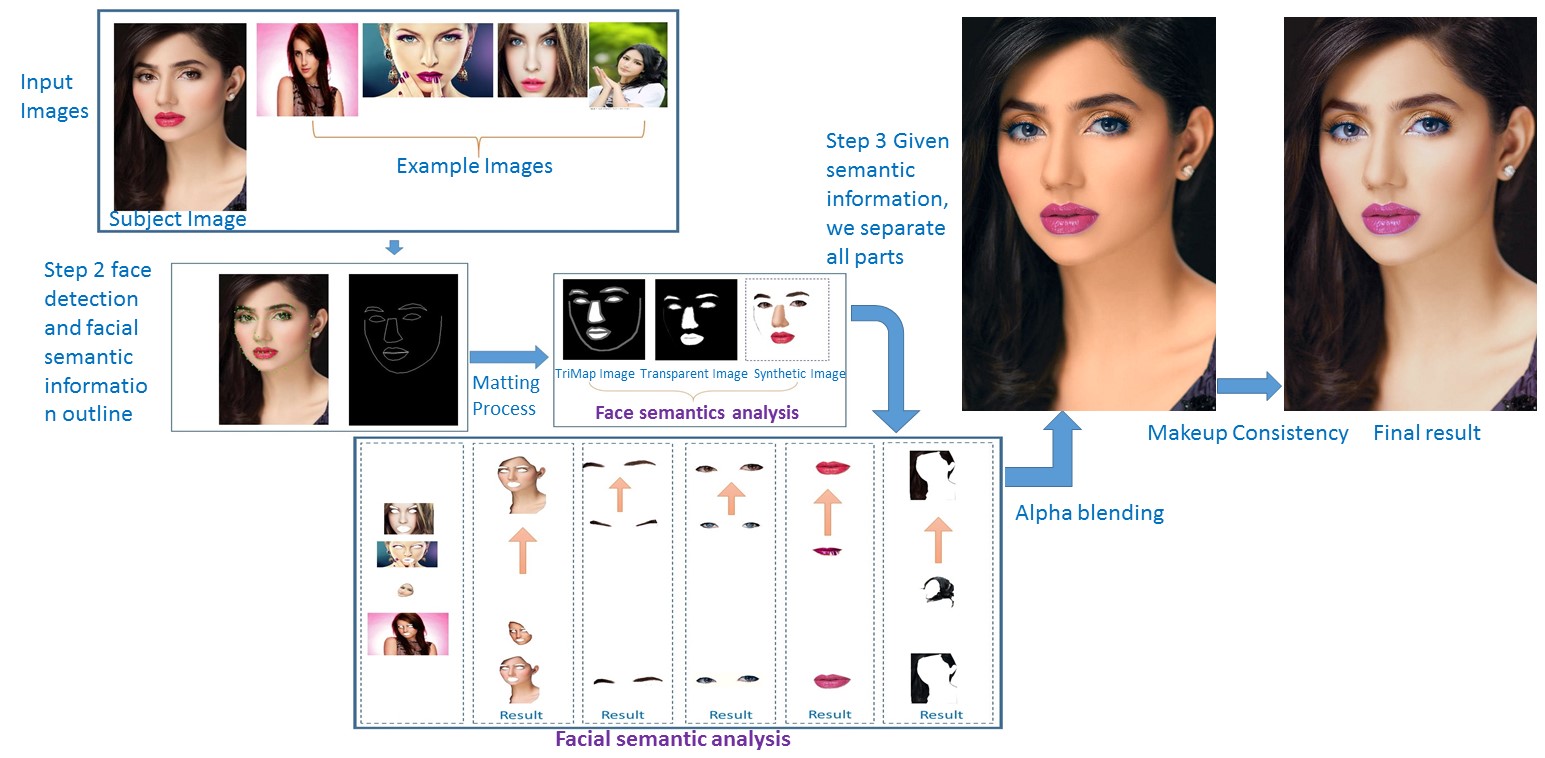}
  \caption{In step 1, we specify the subject and example images, in step 2, we take semantic information after face detection,
          in step 3, we extract the semantic information by using matting algorithm, in step 4, we separate all parts of given sematic information,
          in fifth step, we get resulting image by using alpha blending and in the final step, we apply makeup consistency to obtain our result with optimized color.}
  \label{fig:pipeline}
\end{figure*}
\noindent\textbf{3.3.    Matting of Face Semantics}\\

A commonly used approach to extract semantic information is the Mean-Shift image segmentation
technique \cite{18}. However, it will produce unwanted hard boundaries between
semantic regions. We employ the image matting technique to obtain semitransparent
soft boundaries. Here we implement our automatic matting based on their matting technique by taking
advantage of our generated TriMap. Existing natural matting algorithms often require a user to identify background, foreground,
and unknown regions using a manual segmentation. However, constructing a suitable TriMap is very tedious and time-consuming.
Sometimes, the inaccuracy in TriMap will lead to a poor matting result.\\

In order to solve the problem mentioned above, we expand the contour of facial semantic
information using an expansion algorithm. After distinguishing the foreground, background
and unknown region by different colors (we set foreground to white, background to black,
and the unknown region to gray), we can obtain a corresponding TriMap for an image.
The mathematical expression for the expansion algorithm is:
\begin{equation}
It(x,y)=\max\limits_{(x',y'):element((x',y'))\neq 0}Is\big(x+x',y+y'\big)
\end{equation}
Consequently, the matting image is computed with our automatically generated TriMap Chen et al.\cite{22}.
The transparent image and the synthetic image are shown in Fig. \ref{fig:trimap}.
The automatic matting approach is also applied in subject images to obtain the basic
semantic segmentation.\\

\noindent\textbf{3.4.    Semantics Makeup Style Transfer}\\

The first step in our
makeup transfer approach is to run white-balancing on both the
subject and the example images. The next step is to match the overall brightness between the
two images. We use the transformed luminance values for
this step and adopt Nguyen et al. \cite{23}.
This technique was unique in its consideration of the scene illumination and the constraint
that the mapped image must be within the makeup style gamut of the example image. The mathematical equation is
\begin{equation}
L_i = C^{-1}_e(C_s(L_s)),
\end{equation}
where $L_s$, $L_i$ and $L_e$ are the subject luminance ,intermediate luminance and example luminance respectively.
$C_s$ and $C_e$ are the cumulative histogram of $L_s$ and $L_e$
respectively. Next, the output luminance $L_o$ is obtained by solving the following linear equation.
\begin{equation}
[I+\sigma(G_x\ ^TG_x+G_y\ ^TG_y)]L_0 =L_i+\sigma(G_x\ ^TG_x+G_y\ ^TG_y)L_s,
\end{equation}
where $I$ is the identity matrix; $G_x$, $G_y$ are two gradient matrices
along x, and y direction; $\sigma$ is a regularization parameter.\\

\begin{figure*}
  \centering
  \includegraphics[width=\linewidth]{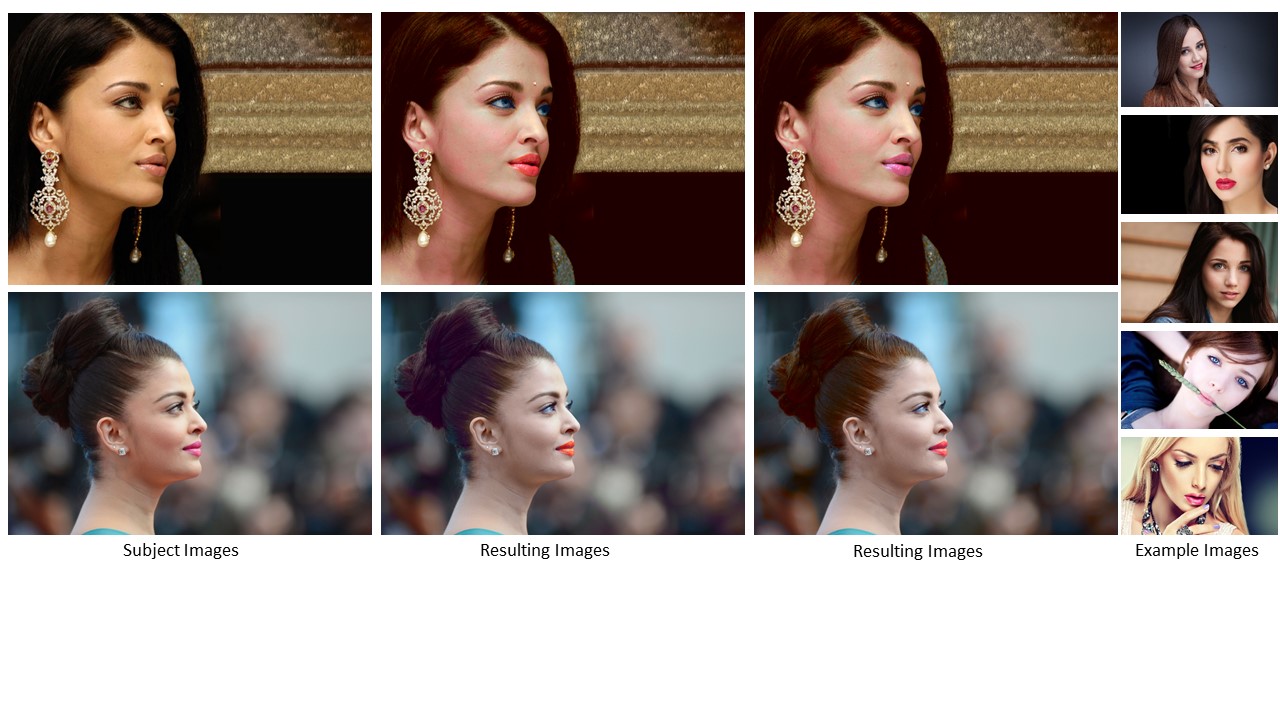}\vspace{-1.8cm}
  \caption{Some results of side pose where the example images are not needed to be of same pose or of same style. The example images are not specific to different results. The texture in the background of the result in first row is preserved, also the background of the result in second row is preserved where it comprises a combination of colors.}
  \label{fig:face pose}
\end{figure*}

To align the subject makeup style gamuts to the example resulting from
the previous step, the centers of the subject and the example image
gamuts are estimated based on the mean values $\mu_s$ and $ \mu_e$
of the subject and example images.
\begin{equation}
\begin{split}
I_s =I_s-\mu_s,\\
I_e=I_e-\mu_e.
\end{split}
\end{equation}

Given a subject and example image, we can propagate makeup style by
minimizing the following energy
\begin{equation}
E = 2\eta((E\times D_s)\oplus D_e )-\eta D_e -\eta(E\times D_s),
\end{equation}
where $D_s$, and $D_e$ are the full 3D convex hulls of the
subject and example image respectively. The operator $\oplus$ is the
point concatenation operation between two convex hulls and
the operator $\eta$ is the volume of the convex hull. A volume
of a combination of two convex hulls is always larger
or equal to that of individual convex hull.\\

\noindent\textbf{3.5.    Makeup Consistency Optimization}\\

We adopt a globally makeup correction model for reasons
discussed in \cite{17,19}, namely robustness to alignment errors,
ease of regularization and higher efficiency due to fewer unknown
parameters. Our simple model is as follows:
\begin{equation}
I^{\prime}=(aI)^\gamma
\end{equation}
where $I^{\prime}$ is the input image, $I$ is the desired image, $a$ is a
scale factor equivalent to the white balance function \cite{20}
and $(.)\gamma$ is the non-linear gamma mapping.
\begin{equation}\label{eq7}
I_i(x_{ij})= (a_ik_jv_{ij})^\gamma{i}
\end{equation}
where $k_j$ is the constant albedo of the $j-th$ 3D point and
$a_i$ and $\gamma_i$ are the unknown global parameters for the $i-th$
image. The per-pixel error term denoted as $v_{ij}$ captures un modeled
color variation due to factors such as lighting and
shading change.\\

Taking logarithms on both side of Eq. \ref{eq7}, and Rewriting in matrix form, by grouping image intensities
by scene point into sparse column vectors of length $m$
and stacking the $n$ columns side by side, we get:
\begin{equation}
I = A+K+V.
\end{equation}

Here, $n$ denotes the number of 3D points or equivalently
the number of correspondence sets. $I\in R^{m\times n}$ is the
observation matrix, where each entry $I_{ij} = log(I_i(x_{ij}))$.
$A\in R^{m\times n}$ is the color coefficient matrix where $A_{ij} = \gamma_{ij} log a_{ij}$ .
$K\in R^{m\times n}$ is the albedo matrix where $K_{ij} = \gamma_{ij}log k_{ij}$ .
Finally, $V\in R^{m\times n}$is the residual matrix where
$V_{ij} = \gamma_{ij}log v_{ij}$ . Here, the row index $i$ denotes the $i-th$
image, and the column index $j$ denotes the $j-th$ 3D point.
\begin{figure*}
  \centering
  \includegraphics[width=0.9\textwidth, height=0.95\textheight]{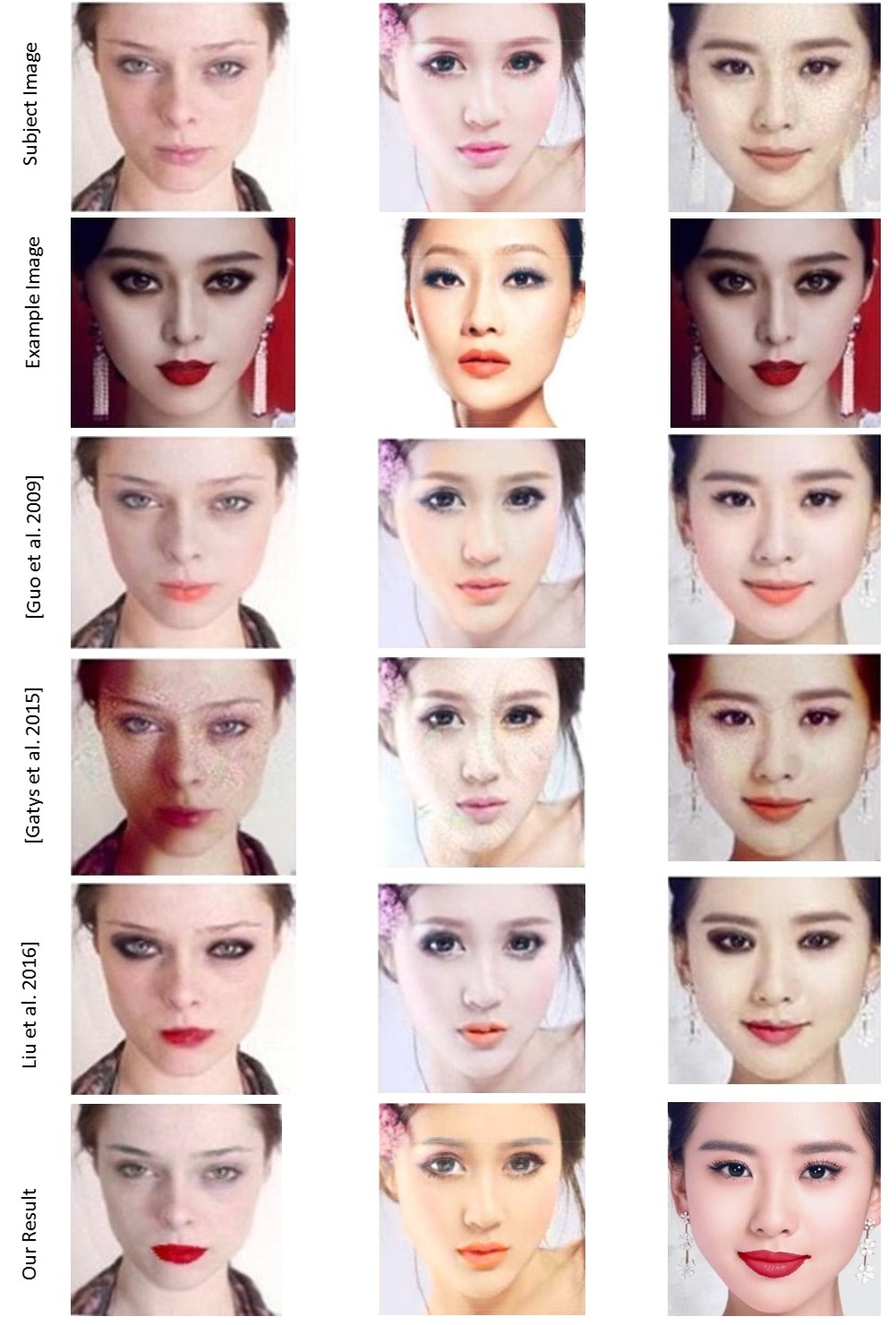}
  \caption{Qualitative comparisons between the state-of-the-arts and ours.}
  \label{comp:makeup}
\end{figure*}

\section{Results and Discussion}

In our method, one can use more than one example images which makes it significant enough.
As sometimes the user likes some parts of the face of one image and other parts of some other image,
our technique allows the user to use different example images. This gives the more natural results
with better visual effects. The semantic analysis of the subject image by using our
interactive tool takes about 2 seconds, and the makeup style transfer step takes about 1 second
with Matlab2014a on a PC with an Intel(R) Core (TM) i5-4690 CPU, 3.50 GHz processor and 8GB RAM under Windows OS.
In Fig. \ref{fig:pipeline}, the pipeline to our technique is exhibited while explaining the
different steps in a sequence.  Our method comprises simple and user-friendly steps to get completed and produces the eye-catching
results with better visual effects. Once we specify the subject image with a number of example images in first step, we take
semantic information after face detection which is automatically done by our algorithm. In next step, it makes use of the
matting algorithm after extracting the semantic information. In second last step, we get resulting image while using the alpha blending
and in the final step, we obtain our final result with optimized makeup after applying makeup consistency.\\

Furthermore, our main focus is inside-eye rather than eye shadow
which produces more natural results. As it is contemporary to use lenses inside the eyes to change
colors of inside-eye, our method focuses entirely on transferring eye style inside eyes and on hairs while preserving
the boundary effectively. It is also important to remark that all the existing techniques do not employ makeup consistency and
use single example image. We focus on using multiple example image rather than single image and employ makeup consistency which
significantly enhances the practicability of our method.
Consequently, we obtain more natural contemporary results as compared with already existing
techniques which only focuses on lips and eye shadows without proper boundary preservation.\\

The qualitative results are shown in Fig. \ref{comp:makeup}.
The first column shows the subject images whereas the second
column shows the corresponding example images. The results by Guo et al. \cite{1},
Gatys et al. \cite{13}, Liu et al. \cite{27} and ours are listed in the third, fourth, fifth and
last columns respectively. First we discuss the comparison with Guo et al. \cite{1}.
Although the results by Guo et al. \cite{1} looks natural but there are number of gaps
which should be removed. For example, Guo et al. \cite{1} always transfers much lighter makeup
then the example face. Since the lip gloss in the example is dark red whereas the lip gloss
by Guo et al. \cite{1} is orange-red. Our method produces more natural results where the lip gloss is
very close to dark-red in the example image. Moreover,
Guo et al. \cite{1} can only produce very light eye shadow even though their focus is eye shadows.\\

Compared with Gatys et al. \cite{13}, our subject faces contain much less artifacts.
It is because our makeup transfer is conducted between the local
regions, such as lip vs lip gloss while the Gatys et al. \cite{13} transfer the makeup
globally. Global makeup transfer suffers from the mismatch
problem between the image pair which clearly shows the disadvantage of the
method by Gatys et al. \cite{13}.\\

Liu et al. \cite{27} focuses on eye shadows and lips whereas skin color remains unaffected.
But the eyes shadows of both eyes are not same in the resulting images which clearly is
a mismatch and records a disadvantage. Moreover, the boundary on the lips and eyes is
not preserved. Our method focuses on inside-eyes, lips, skin
and hairs while keeping the boundary well-preserved. This produces more attractive results of
exquisite nature.\\

Thus our method clearly fills the gaps
which were limitations to the existing techniques.\\

\begin{figure*}
  \centering
  \includegraphics[width=1.0\textwidth, height=0.97\textheight]{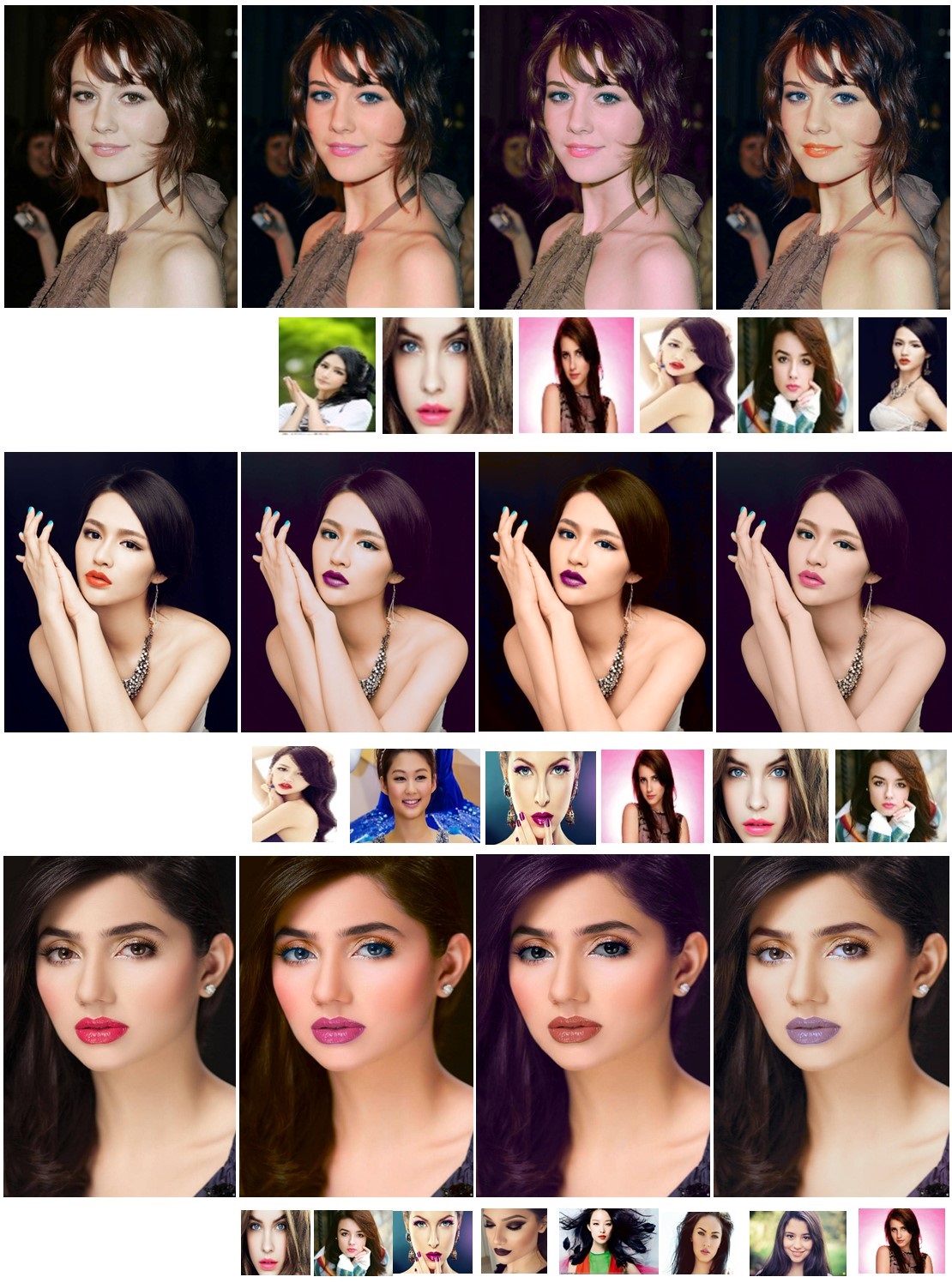}
  \caption{A result with the subject image in the leftmost column and resulting images in second, third and fourth columns with respective example images.}
  \label{fig:final result}
\end{figure*}
\begin{figure*}
  \centering
  \includegraphics[width=\linewidth, height=0.97\textheight]{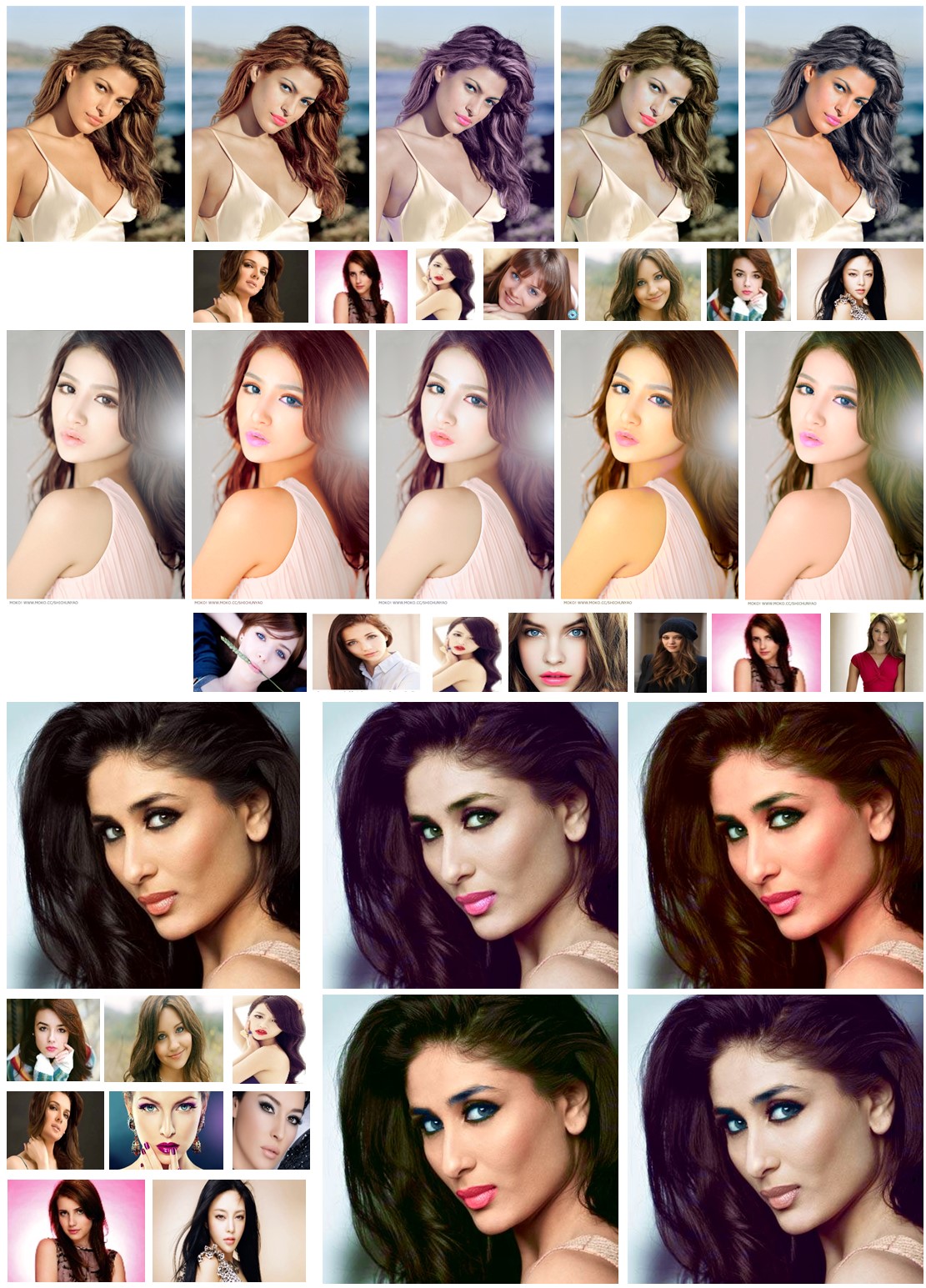}
  \caption{Multi-pronged application of our method,}
  \label{fig:further result}
\end{figure*}

\noindent\textbf{4.1.    More Results On Makeup Transfer}\\

In Fig. \ref{fig:image1}, an application of our method, on a subject image while considering a number
of example images, is presented. We consider four example images to transfer the makeup style in
a subject image and the resulting image shows the result after application of our technique.
It is not customary to use multiple example images in earlier work on this topic. Nevertheless,
we make use of the multiple example images which enhances the efficiency of this work. In Fig. \ref{fig:different style},
a number of results of efficient makeup transfer with different facial styles are presented. A single subject image is
considered and a variety of results, while considering four example images, are presented which shows the diversity
of our technique.\\

In Fig. \ref{fig:face pose}, we apply our algorithm to some images with side pose depiction. We see that our technique is also
suitable for side pose images and it produces the results of same quality as for other images. Moreover, the texture from the background of the
result in first row is preserved well and the background of the result in second row where it is a combination of different colors is also preserved
efficiently. The main purpose of this experiment is to extend the limitation of our method from normal images to side pose images. We show that
our method works for these images types as well and produce the results of same quality.\\

In Fig. \ref{fig:final result}, some results of our technique are presented. As our method is not restricted to one example image only, the
results are given with multiple example images. Our technique transfers the makeup of different parts of the face like hairs, lips and eyes etc.
in the subject image by taking semantic information of similar nature from multiple example images. Our technique transforms an image of low characteristics
to an artistic and exquisite image in just a few steps. The results in Fig. \ref{fig:final result} shows the efficiency of our method as the results are
relatively better than of existing techniques.\\

In Fig. \ref{fig:further result}, shows some more results of our proposed method using multiple example images.
They all show the makeup-transferred results that reflect the example colors to the subject images effectively.
Moreover, the makeup preservation in the resulting image is focused and tackled successfully.
We consider images with different poses and show the effective applicability of our method on these image types.
Fig.  \ref{fig:further result}, shows that our method can successfully transfer the makeup in subject images with different
styles and types e.g. side pose, back pose and other poses which are normally tough to tackle.
One can choose a suitable combination of makeups which seem more attractive with respect to different poses and styles.\\

\begin{figure}[h!]
  \centering
  \includegraphics[width=\linewidth]{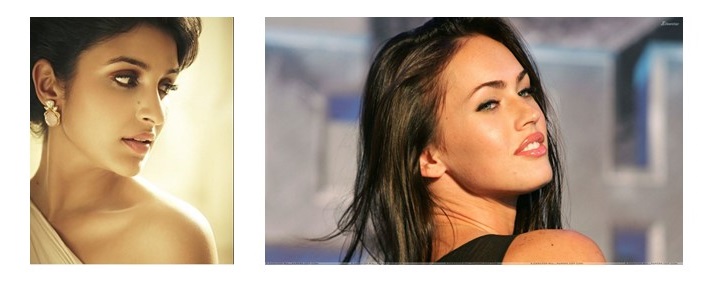}\vspace{-0.5cm}
  \caption{A drawback to our method: The image in first column shows a
failure due to lack of semantic constraints and the image in second column
shows problem caused by lack of clarity of face information.}
  \label{limitation}
\end{figure}

\noindent\textbf{Limitation:} The limitations of our technique are exhibited in Fig. \ref{limitation}.
In the image on left side, the face skin color matches with the background color, therefore
our algorithm does not detect the face and thus unable to extract the semantic information.
In the image on right side, the face information is not clear and therefore leads to a failure
where the algorithm is not able to detect the face and hence lacks extracting the facial semantic
information.

\section{Conclusion}

We have presented a system for stylizing possible appearances
of a person from a single or more photographs.
We have proposed a new semantic makeup transfer technique between images to efficiently change the
makeup style of images. In just few steps, our proposed framework can transform a common image of low characteristics
to an exquisite and artistic photo. The user is not required to select subject and example images with face matching,
as our method can make use of the multiple example images. The broad area of
applications of our technique includes high level facial semantic information and scene content analysis
to transfer the makeup between images efficiently.
Moreover, our technique is not restricted to head shot images as we can also change the makeup style in the wild.
The advantage of using multiple example images is to choose your favorite makeup style from different images as you are
not restricted to choose the makeup style from a single image.
A number of results are presented in different styles and conditions which shows the ubiquitousness and diversity of our method
to industrial scale.
While minimizing manual labor and
avoiding the time-consuming operation of semantic segmentation for the example image,
the framework of our proposed method can be broadly used in film post-production, video-editing, art and design and image processing.

\section{References}

\bibliographystyle{model3-num-names}
\bibliography{<your-bib-database>}

\end{document}